\pdfoutput=1

\documentclass[11pt]{article}

\usepackage[preprint]{acl}

\usepackage{times}
\usepackage{latexsym}

\usepackage[T1]{fontenc}

\usepackage[utf8]{inputenc}

\usepackage{microtype}

\usepackage{inconsolata}

\usepackage{graphicx}
\usepackage{xcolor}
\usepackage{xspace}
\usepackage{amsmath}
\usepackage{cleveref}
\usepackage{framed}
\usepackage{mdframed}
\usepackage{multirow}
\usepackage{booktabs}
\usepackage{pifont}
\usepackage{fdsymbol}
\usepackage{tabularx}

\newcommand{\method}{\textsc{PB\&J}\xspace}
\newcommand{\opinionqa}{OpinionQA\xspace}
\newcommand{\movielens}{MovieLens\xspace}

\newcommand{\nopersona}{\textsc{No Persona}\xspace}
\newcommand{\onlydemo}{\textsc{Only Demographics}\xspace}
\newcommand{\onlyjudge}{\textsc{Only Judgments}\xspace}
\newcommand{\hwang}{\textsc{Demographics + Judgments}\xspace}

\newcommand{\noscaffold}{\textsc{No Scaffold}\xspace}
\newcommand{\experiences}{\textsc{Experiences}\xspace}
\newcommand{\bigfive}{\textsc{Big 5 Personality Traits}\xspace}
\newcommand{\shwartz}{\textsc{Schwartz Theory of Basic Human Values}\xspace}
\newcommand{\primalbeliefs}{\textsc{Primal World Beliefs}\xspace}
\newcommand{\humanwritten}{\textsc{Human Written}\xspace}
\newcommand{\gptfour}{GPT-4\xspace}
\newcommand{\mistral}{Mistral 7B\xspace}

\newcommand{\eg}{e.g.\xspace}

\title{Improving Language Model Personas via Rationalization\\ with Psychological Scaffolds}

\DeclareSymbolFont{extraup}{U}{zavm}{m}{n}
\DeclareMathSymbol{\vardiamond}{\mathalpha}{extraup}{87}
\DeclareMathSymbol{\varheart}{\mathalpha}{extraup}{86}
\newcommand{\aspace}{\hspace{0.25em}}
\newcommand{\apple}{$^{\vardiamond}$}
\newcommand{\usc}{$^{\varheart}$}

\author{
    Brihi Joshi\usc\aspace 
    \textbf{Xiang Ren}\usc\aspace
    \textbf{Swabha Swayamdipta}\usc\aspace
    \textbf{Rik Koncel-Kedziorski}\apple\aspace
    \textbf{Tim Paek}\apple\aspace\\
    \usc University of Southern California \aspace
    \apple Apple \\
    \texttt{\{brihijos\}@usc.edu}
}

\begin{document}
\maketitle
\begin{abstract}
Language models prompted with a user description or {\it persona} are being used to predict the user's preferences and opinions.
However, existing approaches to building personas mostly rely on a user's demographic attributes and/or prior judgments, but not on any underlying \textit{reasoning} behind a user's judgments. 
We introduce \method (\textbf{P}sychology of \textbf{B}ehavior and \textbf{J}udgments), a framework that improves LM personas by incorporating potential rationales for \textit{why} the user could have made a certain judgment. 
Our rationales are generated by a language model to explicitly reason about a user's behavior on the basis of their experiences, personality traits, or beliefs. 
Our method employs \textit{psychological scaffolds}: structured frameworks such as the Big 5 Personality Traits or Primal World Beliefs to help ground the generated rationales in existing theories. 
Experiments on public opinion and movie preference prediction tasks demonstrate that language model personas augmented with \method rationales consistently outperform personas conditioned only on user demographics and / or judgments, including those that use a model's default chain-of-thought, which is not grounded in psychological theories.
Additionally, our \method personas perform competitively with those using human-written rationales, suggesting the potential of synthetic rationales guided by existing theories.

\end{abstract}

\section{Introduction}
\label{sec:introduction}

\begin{figure*}[!h]
    \centering
    \includegraphics[width=\linewidth]{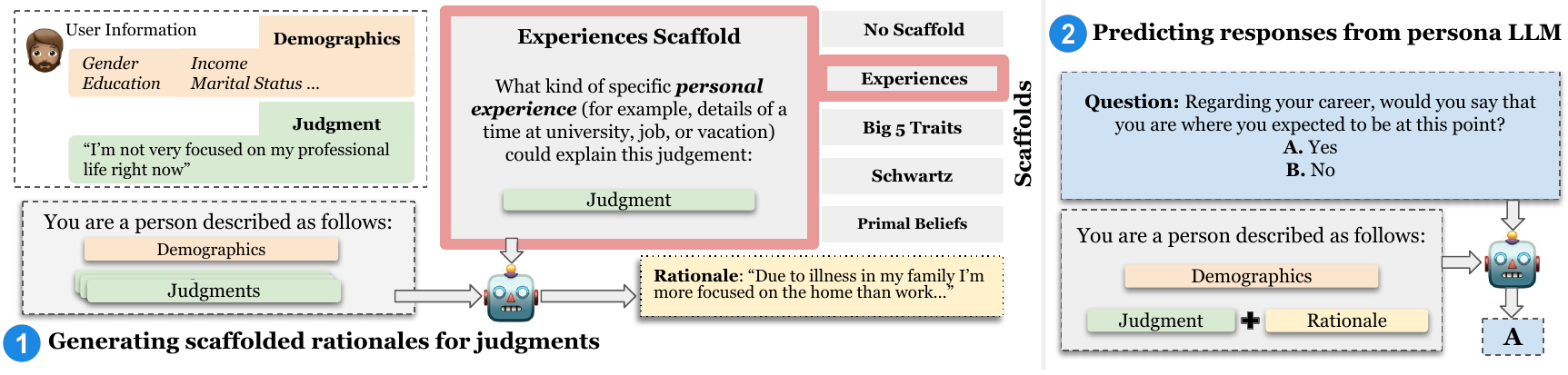}
    \caption{\textbf{Overview of the \method framework:} A base persona comprising user Demographics and Judgments is augmented with post-hoc rationales of various Scaffolds generated by an LM. The updated persona is integrated into the system prompt of an LM to align predictions with user behavior. 
    }
    \label{fig:method_figure}
\end{figure*}

Recent advances in language modeling for user \textit{persona} (i.e, characteristics, preferences, and behavior of a user) offer straightforward yet powerful ways to predict user behavior and decision-making \cite{DOnofrio2020personae}.
Personas have enabled simulation capabilities in LMs, such as helping survey design by simulating a wide range of human responses \cite{Argyle_2023, santurkar2023opinionslanguagemodelsreflect, Tjuatja2023DoLE}, simulating communities to study behavior \cite{park2022socialsimulacracreatingpopulated, zhou2024sotopiainteractiveevaluationsocial,park2024generativeagentsimulations1000}, and producing diverse and large-scale synthetic data \cite{moon2024virtualpersonaslanguagemodels, ge2024scalingsyntheticdatacreation}.
Yet, simulated LM personas still struggle to align well with intended user behavior 
\cite{gupta2024biasrunsdeepimplicit, liu2024evaluatinglargelanguagemodel}.

A typical task for evaluating LM personas is opinion prediction: how accurately does the LM persona reflect the user's real opinion to questions such as
``\textit{For your job or career aspects of your life, would you say that you are where you expected to be at this point in your life?}''
Fine-tuning LMs on user conversations and interaction history to create personas does not scale well due to the lack of adequate user-specific data \cite{mazare-etal-2018-training,madotto-etal-2019-personalizing,li-etal-2024-steerability}.
Zero-shot prompting approaches, where a persona is specified using demographic attributes in a system prompt, avoid these issues but often produce inconsistent responses and expose known biases in current LMs~\cite{santurkar2023opinionslanguagemodelsreflect,hu-collier-2024-quantifying,cheng-etal-2023-marked, gupta2024biasrunsdeepimplicit}.
To improve zero-shot prompting, recent approaches incorporate user \textit{judgments} (prior user responses and history) in the prompt \cite{hwang-etal-2023-aligning, sun2024personadbefficientlargelanguage}.
For example, a user judgment like ``\textit{I'm not very focused on my professional life right now}''
could inform their response to questions about their career expectations.
While these methods provide additional context, they still fail to address a critical gap: the ability to \textit{rationalize why} a user might have a specific judgment, an essential component to understand behavior.
In our example, \textit{one plausible rationale} for the user's judgment
could be that they are prioritizing their family over their career at this time. 

In this work, we hypothesize that incorporating \textit{plausible post-hoc rationales for a user's judgment} could improve LM personalization, bridging the gap between surface-level personas and deeper, more nuanced simulated behavior.
However, collecting such rationales is expensive and existing datasets for building personas only contain judgments \cite{kirk2024prism, durmus2024measuringrepresentationsubjectiveglobal}, and in rare cases, user demographics \cite{santurkar2023opinionslanguagemodelsreflect, harper2015movielens}, but no rationales for judgments.
To address this limitation, we introduce \method (\textbf{P}sychology of \textbf{B}ehavior and \textbf{J}udgments), a framework that uses post-hoc plausible, yet synthetic, LM-generated rationales for user judgments, to improve LM personas.
The rationales for user judgments in our method stem from psychological theories, such as personality traits (e.g., Big 5 traits \cite{goldberg1993bigfive}, Schwartz’s value theory \cite{shwartz1992test}) and belief systems (e.g., Primal World Beliefs \cite{clifton2019primalworldbeliefs}); we call these \textit{psychological scaffolds}.
Each psychological scaffold offers additional context to produce one potential (albeit generated) path of reasoning to support a given user judgment. 
This is in contrast to chain-of-thought rationales \cite{wei2023chainofthoughtpromptingelicitsreasoning} where no explicit psychological scaffold is employed; the LM relies on its own world knowledge to rationalize a test input prediction ad hoc.

It is important to note that while our synthetic rationales may not reflect the \textit{true} reasoning behind a user's judgment, they may contain plausible markers of real behavioral reasoning due to the presence of such reasoning in the LM's training data \cite{binz2023using, hagendorff2023machine, balepur2025boatdoesfloatimproving}.
Indeed, humans use everyday reasoning to rationalize behavior; this is known as {\it folk psychology} \cite{churchland1988folk, malle1997folk, Malle_2004}.


Given a user's demographics and prior judgments, \method produces LM-generated rationales for user's judgments to construct a richer, more comprehensive user description for persona prompting.
Our experiments show that when prompted with such richer descriptions, LMs can result in more accurate personas, as evaluated on a set of test instances. 
We evaluate \method on two tasks that represent different aspects of personalization: opinion prediction in \opinionqa \cite{santurkar2023opinionslanguagemodelsreflect} and movie preference prediction in \movielens \cite{harper2015movielens}. 
On both benchmarks, we show that LM personas prompted with \method rationales are significantly better at predicting a user's response to test instances than those prompted with user demographics and/or prior judgments alone, including with additional chain-of-thought reasoning.
Improvements from \method are seen across nearly all demographic categories, and \method can improve personas with or without demographic information. 
Finally, we present a pilot experiment where we collect human-written rationales, along with their demographics and judgments on a subset of \opinionqa; LM-rationales nearly match the performance of human-written rationales in \method, showing the efficacy of generated rationales in our approach.

At a high level, our work highlights the value of synthesizing data with the appropriate structural context to produce plausible rationalization that could improve the modeling of user opinions and preferences in already capable language models. 
\section{Background}

Humans use common explanations to rationale each other's actions \cite{Malle_2004}.
We hypothesize that rationales are common enough in our discourse (e.g., narrative text of novels, biographies, social media)  to be prevalent in training corpora for LMs to generate effective rationales in zero-shot settings \cite{binz2023using,hagendorff2023machine}.
In order to better guide LMs, we employ various \textit{psychological scaffolds}, which are templates based on more structured theories of psychology, to rationalize user judgment. 

\subsection{Rationalization}
\label{sec:psych:rationales}
Psychologists have long studied how people use everyday language to rationalize and anticipate behavioral patterns in social interactions, also known as folk psychology \cite{churchland1988folk, malle1997folk, Malle2009-MALFTO-3, Gordon1986-GORFPA-2}. 
Building on the example introduced in \Cref{sec:introduction}, suppose that we have to rationalize \textit{why} a user believes that \textit{they are not focused on their professional life and career at the moment} (See \Cref{fig:method_figure}).
We could assume that this behavior arises due to some belief, such as \textit{life is dull and uninteresting and chasing career progressions is a waste of time} \cite{Dennett1981-DENTIS}.
Or we may offer life experiences that shape beliefs, such as \textit{personal life issues that get prioritized over one's career at the moment} \cite{Malle_2004}.
For our purposes, we define \textbf{rationalization} as an attempt to understand the behavior of others using explanations based on assumed
mental states. 


Unlike prior work, which uses rationalization as reasoning chains to explain model decisions ad hoc, to improve model task performance \cite{wei2023chainofthoughtpromptingelicitsreasoning, zelikman2022starbootstrappingreasoningreasoning, ramnath2024tailoringselfrationalizersmultirewarddistillation}, or to enhance human utility \cite{joshi-etal-2023-machine, si2024largelanguagemodelshelp, chaleshtori2024evaluatingexplanationutilityhumanai}, we focus on post-hoc rationalization of user judgments to improve LM personas for accurate user predictions and preferences. 
Formally, given a user persona $\mathcal{Q}$
and judgment $j$ as input,
we instruct an LM to generate ``a reasonable explanation that the user would provide for holding that judgment.''
The {\it basic rationale} $r_\text{basic}$ is defined as 
\begin{equation}
\label{eqn:independent_rationalization}
    r_{\text{basic}} = LM_{\mathcal{R}}(j, \mathcal{Q}),
\end{equation}
where $LM$ is an off-the-shelf pretrained language model instructed to elicit a rationale.
\subsection{Psychological scaffolds}
\label{sec:psych:scaffold}
\begin{table}[!t]
\centering
\resizebox{\columnwidth}{!}{%
\begin{tabular}{p{3cm}p{6cm}}
\hline
\textbf{Scaffold Name} & \textbf{Description} \\ \hline
\noscaffold & Free-form rationales without any structured guidance which rely on the model's internal knowledge of folk psychology. \\ \hline
\experiences \cite{mcadams1993stories} & Rationales describing the life experiences and societal norms that would motivate a user's judgment, such as relationships, work, or community. \\ \hline
\bigfive \cite{goldberg1993bigfive} & Rationales use personality traits such as openness, conscientiousness, extraversion, agreeableness, and neuroticism to explain judgments. \\ \hline
\shwartz \cite{shwartz1992test} &  Rationales that ground judgments in the user's relationship to stimulation, hedonism, self-direction, universalism, and security, among others. \\ \hline
\primalbeliefs \cite{clifton2019primalworldbeliefs} & Rationales which use worldview --- whether the world is good, safe, alive, or enticing --- to explain user judgments. \\ \hline
\end{tabular}%
}
\caption{Summary of the scaffolds used in \method, highlighting their theoretical underpinnings and practical relevance.}
\label{table:rationale_scaffolds}
\end{table}

While \Cref{eqn:independent_rationalization} helps us generate rationales from LMs, it is important to control \textit{what} LMs generate as plausible rationales for human behavior. 
Can LMs generate more informed rationales based on well-established psychological theories? 
To find out, we employ \textit{psychological scaffolds} to enhance the LMs' reasoning capabilities about user beliefs and preferences by providing coherent scaffolds as instructions to follow for reasoning about the underlying causes of observed behavior. 

The specific scaffolds we study in this work are shown in \Cref{table:rationale_scaffolds}.
In the context of our previous example, the \textsc{experiences} scaffold might promote a rationale based on the bad experience of a family member who prioritized career goals over personal life \cite{mcadams1993stories}.
Rationales generated from the \textsc{personality traits} scaffold may posit the user's openness to new experiences, including bold career choices \cite{goldberg1993bigfive, shwartz1992test}.
A rationale using the \textsc{Primal World Beliefs} scaffold may surmise that the user sees the world as dull and uninteresting and thus not worth the effort of seeking professional success \cite{clifton2019primalworldbeliefs}.
More details about the theories underlying each scaffold can be found in \Cref{sec:apx:psych_scaffolds}.
While the scaffolds that we select in this work are common psychological constructs used in prior work on personality and social intelligence~\cite{zhou2024sotopiainteractiveevaluationsocial, vu2022modeling, moon2024virtualpersonaslanguagemodels}, any scaffold that helps reason about a user's belief can be used. 
Our selected scaffolds offer diverse yet complementary ways to structure psychological rationales, anchoring rationales in personal experiences, personality traits, and belief systems. 

To generate a rationale from an LM with a particular scaffold, we include additional scaffold information in the rationale generation instructions to arrive at scaffold-specific instructions $\psi$. Using
$\mathcal{Q}$ and $j$ as above, the {\it scaffold rationale} $r_\psi$ is defined as follows:
\begin{equation}
\label{eqn:scaffold_rationalization}
    r_{\psi} = LM_{\mathcal{R}}(j, \mathcal{Q}, \psi)
\end{equation}
The specific instructions used for each scaffold are given in Appendix~\ref{sec:apx_prompts}.
\section{LM Personas using \method}
\label{sec:method}

Consider a task where an LM has to adopt the persona of a given user, and perform a task (\eg answer opinion-based questions) as their proxy.
In this section, we describe how we leverage LM rationalization
to construct LM personas, which are used as prompts to generate personalized responses as a proxy for the user.
We begin by setting up a basic persona description composed of user demographics and augmented with a set of user judgments, just as in previous work \cite{hwang-etal-2023-aligning}.
We then present \method, which enriches persona descriptions by providing additional context in the form of rationales, and demonstrate how \method is used to generate personalized responses as a proxy for the user.

\begin{figure}[!t]
\small

\begin{mdframed}[linecolor=black!30,backgroundcolor=black!5]
\textbf{OpinionQA}\\
\textit{Question}: How well, if at all, do the following words or phrases describe you? Interested in visiting other countries?\\
\textit{Options}:\\
A. Describes me well\\
B. Does not describe me well\\
C. Refused to answer\\
\textit{User-selected Answer}: A. Describes me well
\end{mdframed}

\begin{mdframed}[linecolor=black!30,backgroundcolor=red!5]
\textbf{MovieLens}\\
\textit{Question:} Out of 5, what would 'To Kill a Mockingbird (1962)' be rated?\\
Synopsis: In small-town Alabama in 1932, Atticus Finch (Gregory Peck) is a lawyer and a widower. He has two young children, Jem and Scout. Atticus Finch is currently defending Tom Robinson, a Black man accused of raping a white woman. Meanwhile, Jem and Scout are intrigued by their neighbors, the Radleys, in particular the mysterious, seldom-seen Boo Radley. \\
Directors: Robert Mulligan\\
Cast: Gregory Peck, John Megna, Frank Overton, Rosemary Murphy, Ruth White\\
\textit{User-selected Answer}: I would rate 'To Kill a Mockingbird (1962)' a 2 out of 5.
\end{mdframed}

\caption{\textbf{Example Task:} Shown here are example inputs and outputs for \opinionqa and \movielens datasets respectively. Each of these instances have corresponding user selected answers.}
\label{fig:dataset_examples}
\end{figure}

\subsection{Persona Building}
\label{sec:method:building_persona}
\paragraph{Base Persona Setup.}
Following \citet{hwang-etal-2023-aligning}, we begin with a {\it basic persona}  \( \mathcal{Q}_B \) for an arbitrary user
consisting of demographic attributes \( \mathcal{D} \) and a set of \( N \) \textit{seed judgments} \( \mathcal{J} \).  
Demographic attributes \( \mathcal{D} \) capture the sociological characteristics and group identity traits of a user, such as age, gender, education, and race \cite{santurkar2023opinionslanguagemodelsreflect}. 
While these attributes can provide useful information on user preferences and opinions, prior work has shown that relying solely on demographics often results in stereotypical and biased responses from LMs \cite{hwang-etal-2023-aligning, cheng-etal-2023-compost, gupta2024biasrunsdeepimplicit}.  
To address these limitations, \citet{hwang-etal-2023-aligning} augment demographic attributes with a set of seed judgments \( \mathcal{J} \).
Seed judgments often represent user-selected answers from a held out set of questions, or a prior interaction history which aims to provide additional context to enrich the persona description \cite{hwang-etal-2023-aligning, do2024choirecharacterizingpredictinghuman}.
In other tasks, a user's interaction history, such as the movies they watched beforehand along with their reviews, constitute the seed judgments.

\paragraph{Post-hoc Rationales for Seed Judgments.}
In \method, we augment the description of the above persona \( \mathcal{Q}_B \) with additional context in the form of post-hoc
generated rationales $r$ for each seed judgment $j \in \mathcal{J}$ following Equation~\ref{eqn:independent_rationalization}. 
Each post-hoc rationale is generated independently for the corresponding seed judgment and is added to the persona alongside the judgment itself.  
Our {\it rationalized persona} is defined as:  
\begin{equation}
    \mathcal{Q}_{\mathcal{R}} = (\mathcal{D};\{(j, r_{\text{basic}}) : j \in \mathcal{J}).
\end{equation} 

Building on \Cref{sec:psych:rationales}, post-hoc rationales operationalize how $LM_{\mathcal{R}}$ generates rationales to explain \textit{why} a user might hold a given judgment. 

Using rationales from Equation~\ref{eqn:scaffold_rationalization}, we arrive at the {\it scaffold rationalized persona},
\begin{equation}
    \mathcal{Q}_{\mathcal{\psi}} = (\mathcal{D};\{(j, r_{\mathcal{\psi}}) : j \in \mathcal{J}).
\end{equation} 

\subsection{Generating responses from Persona LMs}
\label{sec:method:using_persona}

Persona descriptions \( \mathcal{Q} \) obtained in \S\ref{sec:method:building_persona} can be used as prompts to steer responses towards a given user persona, following prior work \cite{santurkar2023opinionslanguagemodelsreflect, hwang-etal-2023-aligning}. 
Consider a task where an individual user has provided several input-output pairs for a test instance \( \{x, y\} \) (\eg, opinion question and answer pairs).
The persona description can provide additional context to predict responses for inputs, enabling an LM to generate accurate predictions for a user.

We predict a user's output for an input $x$ by prompting an off-the-shelf language model $LM_{\mathcal{T}}$ with task-specific instructions $\mathcal{T}$.
\begin{equation}
    \hat{y} = LM_{\mathcal{T}}(x,\mathcal{Q}),
\end{equation}
To evaluate a persona building method, we vary $\mathcal{Q}$ over the base, rationalized, and scaffold rationalized personas $\mathcal{Q}_B, \mathcal{Q}_\mathcal{R}, \mathcal{Q}_{\psi}$. 

\section{Experiments}
\label{sec:experiment}
\begin{table}[t!]
    \footnotesize
    \centering
    \resizebox{\columnwidth}{!}{
    \begin{tabular}{p{\columnwidth}}
        \toprule
        \textbf{Judgment:} I am somewhat dissatisfied with my current job or career. \\
        \midrule
        \midrule
        
        \textbf{\noscaffold} \\
        At my age, I've had many experiences in my working life and feel that I could have achieved more. While I continue to work because I believe in keeping busy and maintaining a sense of purpose, my current job doesn't give me the satisfaction and fulfillment I'd hoped for...\\

        \\

        \textbf{\experiences} \\
        As a man in my senior years, I've spent decades working. I began in manual labor, but found a calling in community work and completed an associate's degree in social work. While this transition has been rewarding, I've grown somewhat dissatisfied with my job mainly because of ever-increasing paperwork and bureaucracy. They are detracting from the direct, meaningful interactions that I so enjoy and that rekindled my career passion in the first place. As I age, I've found that what I value most in work is not status or power...\\ 
        \\

        \textbf{\bigfive} \\
       This dissatisfaction may stem from extroversion and my desire for more social interaction that my current job may not provide. My conscientiousness also comes into play as I am a productive and organized individual who might feel stifled if these traits are not fully utilized in my career. At the same time, my neuroticism may cause me to overanalyze situations and magnify any negative aspects of my job...\\
        \\

        \textbf{\shwartz} \\
        This opinion reflects my value of Achievement and Self-Direction. I value being successful and competent, and currently, my job is not allowing me to feel that sense of accomplishment. Additionally, I value independent actions and the ability to choose my own path, thus being dissatisfied with my current job also relates back to my sense of autonomy and need for personal fulfilment... \\ 
        \\

        \textbf{\primalbeliefs} \\
        Despite being on the older side, I still have an enticement for experiencing new things and learning more. Sticking to the same old job with monotonous routines doesn't appeal to me because it does not satisfy my curiosity or hunger for novelty. Additionally, my safe world belief also comes into play. I believe in stability and minimal threats but my current job doesn’t provide that comfort and ease that I expect at this point in my life. The constant worries and pressures create a sense of dissatisfaction. Lastly, I also believe in the alive world stance. I want a sense of purpose in what I do. I believe there is a reason for my dissatisfaction, and it's perhaps because this isn't the job I was meant to stay in for the rest of my life...\\
        \bottomrule
    \end{tabular}}
    \caption{Snippets of rationales generated by \gptfour for a career satisfaction judgment according to the scaffolds used in \method. Full examples are shown in \Cref{sec:apx:examples}, along with user demographics}
    \label{tab:scaffold_examples}
\end{table}

\paragraph{Task and Datasets.}
We conduct experiments on two datasets - \opinionqa \cite{santurkar2023opinionslanguagemodelsreflect} and \movielens \cite{harper2015movielens}.
\opinionqa is a collection of American public opinion surveys~\cite{pew_atp}
conducted by PEW Research 
containing user-selected answers 
to multiple-choice questions in 15 different topics ranging from food safety to guns, with rich accompanying demographic attributes.
The task is to
predict a user's answer for a question.
For \opinionqa, we use a subset of 750 users, and 10 test questions for each user.
\movielens contains timestamped movie ratings (between 1-5) corresponding to individual users, with gender, age, occupation and location of each user.
The task is to predict a user's rating for a movie.
We use a subset of 100 users and 10 test movies for each user.
Further details and splits of the dataset are presented in \Cref{sec:apx_dataset}. 
Example inputs-outputs of each of these datasets are shown in \Cref{fig:dataset_examples}.
\paragraph{Seed Judgments.}
Our persona description generates rationales for a set of seed judgments.
For \opinionqa, these seed judgments are a `train' set of questions and users' corresponding answers to them.
The questions and answers are then converted into declarative forms (e.g., If a user responds \textit{No, not really} to \textit{“Are you currently focused on your professional life and career?”}, the declarative form would be \textit{I’m not very focused on my professional life right now}).
For \movielens, we use movie ratings provided by the user with earlier timestamps for predicting ratings by the same user with later timestamps. 
Each judgment consists of a movie, its rating, and a short description of the movie consisting of its plot synopsis, actors, and directors. 
For our main experiments, a fixed set of 8 seed judgments are provided in the persona description.

\paragraph{Evaluation.} We use a `test' set of opinion questions and movies to evaluate the personas.
Since our personas are customized to each real user, we calculate accuracies and standard deviation for both \opinionqa and \movielens, macro-averaged for each user. 
\begin{table*}[!t]
\centering
\resizebox{\linewidth}{!}{
\begin{tabular}{lcccc}
\toprule
\textbf{Prompting Approach} & \multicolumn{2}{c}{\textbf{\opinionqa}} & \multicolumn{2}{c}{\textbf{\movielens}}\\ 
\cline{2-5}
 & \textbf{\gptfour} & \textbf{\mistral} & \textbf{\gptfour} & \textbf{\mistral} \\ \hline
 \nopersona  & 20.57 \small{$\pm$ 15.15} & 32.08 \small{$\pm$ 16.83} & 21.89 \small{$\pm$ 13.01} & 06.30 \small{$\pm$ 9.34}\\ 
   \onlydemo & 45.69 \small{$\pm$ 17.46} & 32.97 \small{$\pm$ 16.46} & \underline{35.30 \small{$\pm$ 18.68}} & 22.90 \small{$\pm$ 14.65}\\
   \onlyjudge & 33.63 \small{$\pm$ 16.76} & 38.40 \small{$\pm$ 16.25} & 32.50 \small{$\pm$ 19.10} &  15.30 \small{$\pm$ 17.97}\\
   \hwang  & \underline{49.63 \small{$\pm$ 16.84}} & \underline{42.17 \small{$\pm$ 16.70}} & 34.80 \small{$\pm$ 17.52} & 21.20 \small{$\pm$ 17.65} \\ 
   $\hwang_{CoT}$  &  49.17 \small{$\pm$ 17.10} & 31.67 \small{$\pm$ 15.59} &  30.20 \small{$\pm$ 18.11} & \underline{24.20 \small{$\pm$ 18.30}}\\ 
   \cline{1-5}
   $\method_{\noscaffold}$  & 53.71* \small{$\pm$ 17.92} & 46.96* \small{$\pm$ 16.44}  & 24.80 \small{$\pm$ 15.20} & 22.40 \small{$\pm$ 14.01}\\
  $\method_{\experiences}$ & 54.12* \small{$\pm$ 17.59} & \textbf{47.61* \small{$\pm$ 16.55}} & 30.50 \small{$\pm$ 16.45} & 22.40 \small{$\pm$ 15.37}\\
  $\method_{\bigfive}$ &  53.59* \small{$\pm$ 17.14} & 46.09* \small{$\pm$ 16.75} & 35.79 \small{$\pm$ 17.84} & 29.30* \small{$\pm$ 16.02}\\
  $\method_{\shwartz}$ & 53.45* \small{$\pm$ 17.17} & 45.00* \small{$\pm$ 16.71} & \textbf{39.89* \small{$\pm$ 17.46}} & 26.70* \small{$\pm$ 17.03}\\
  $\method_{\primalbeliefs}$ & \textbf{54.43* \small{$\pm$ 17.01}} & 45.52* \small{$\pm$ 16.19} & 38.00* \small{$\pm$ 16.91} & \textbf{30.50* \small{$\pm$ 16.15}} \\ 
 \bottomrule
\end{tabular}}
\caption{\textbf{Improved Persona Alignment with \method}: Shown here are accuracy scores and standard deviations macro-averaged across users, for \method with different psychological scaffolds (\noscaffold, \experiences, \bigfive, \shwartz, and \primalbeliefs). \method consistently outperforms baselines, demonstrating the effectiveness of scaffolded rationales in improving persona alignment.
Best performing method is \textbf{bolded} and best performing baseline is \underline{underlined}. * represents results that are significantly better ($p < 0.05$) than the best baseline. Full significance results are in \Cref{sec:apx:significance}. 
}
\label{table:main_results}
\end{table*}

\paragraph{Language Models.} We use two LMs of varying sizes --- \gptfour\footnote{Last accessed on 2 December 2024, \gptfour points to \texttt{gpt-4-0613}.}\cite{openai2024gpt4technicalreport} and Mistral 0.2 Instruct 7B \cite{jiang2023mistral7b}.
All results displayed are based on prompting these LMs, without any fine-tuning.

\paragraph{Baselines.} Following prior work, we experiment with different variants of building persona descriptions as baselines, 
\citet{santurkar2023opinionslanguagemodelsreflect} propose two methods -- one where LMs are prompted without any user information (\nopersona) and with demographics (\onlydemo).
\citet{hwang-etal-2023-aligning} add judgments to the persona descriptions without (\onlyjudge) and with demographics (\hwang). 
For the latter setting, we also include results using Chain-of-thought reasoning \cite{wei2023chainofthoughtpromptingelicitsreasoning} as $\hwang_{CoT}$, where $LM_{\mathcal{T}}$ is prompted to reason about the test question using the persona information in order to arrive at the answer.
This is $LM_{\mathcal{T}}$'s default rationale that is not explicitly grounded in any psychological scaffold.

\paragraph{Existing persona LM baselines are (almost) equivalent.}
\Cref{table:main_results} presents macro-averaged accuracies for various baseline methods across models and datasets.
Without any persona information (\nopersona condition), both \gptfour and \mistral perform significantly worse, highlighting the necessity of incorporating some form of user context. 
The performance drop is particularly stark for \gptfour; due to its safety guardrails, \gptfour abstains frequently from expressing opinions and preferences~\cite{chen2023chatgptsbehaviorchangingtime}.
This leads to responses such as 
\textit{"This is a subjective question and the answer will vary"} or \textit{"As an AI, I don’t have personal experiences"}\footnote{Such behavior does not occur for all questions but has a substantial impact on overall performance.}.
Demographic attributes (\onlydemo) are generally more predictive than judgments alone (\onlyjudge) with \hwang as the best performing baseline, especially for \gptfour.
However, it is interesting to see that providing additional tokens to reason via Chain-of-Thought rationales in $\hwang_{CoT}$ also yields no significant improvements over $\hwang$.

\begin{table}[!t]
\centering
\resizebox{\linewidth}{!}{
\begin{tabular}{lccc}
\toprule
\textbf{Approach}  & \textbf{Accuracy} \\ 
\midrule
\nopersona & 08.61 \small{$\pm$ 05.20} \\ 
\onlydemo  & 24.28 \small{$\pm$ 15.81} \\
\onlyjudge & 21.47 \small{$\pm$ 08.50}\\
\hwang  & \underline{39.42 \small{$\pm$ 11.43}} \\ 
$\hwang_{CoT}$ & 39.30 \small{$\pm$ 11.82}\\ 
\midrule
$\method_{\noscaffold}$  & 44.62* \small{$\pm$ 11.42}\\ 
$\method_{\experiences}$  & 43.76* \small{$\pm$ 11.60} \\
$\method_{\bigfive}$ & 44.61* \small{$\pm$ 10.98}\\
$\method_{\shwartz}$  & 45.33* \small{$\pm$ 11.50}\\
$\method_{\primalbeliefs}$ & 46.71* \small{$\pm$ 11.52} \\ 
$\method_{\humanwritten}$  & \textbf{48.52* \small{$\pm$ 12.30}} \\ 
\bottomrule
\end{tabular}}
\caption{\textbf{Incorporating human-written rationales in \method:} Human-written rationales for \opinionqa judgments consistently outperform baselines and all LM-generated rationales except \primalbeliefs.}
\label{table:human_pilot_results}
\end{table}

\paragraph{\method provides significant improvements for personas.}
Across both datasets (\Cref{table:main_results}), \method consistently outperforms the baselines, demonstrating the effectiveness of plausible, synthetic rationales (example generations in \Cref{tab:scaffold_examples}). 
On \opinionqa, scaffolded rationales yield significant improvements over \hwang, even after generating Chain-of-Thought rationales, with \primalbeliefs achieving the highest accuracy, closely followed by \experiences. 
Even unstructured rationales (\noscaffold) surpass $\hwang_{CoT}$. 
These trends hold across LMs.

For \movielens, psychological scaffolds are even more critical. 
While \gptfour performs better overall, smaller LMs like \mistral benefit substantially from \shwartz and \primalbeliefs scaffolds. 
\primalbeliefs consistently ranks among the top scaffolds across datasets, 
however, \shwartz excels in \movielens, highlighting that different tasks may benefit from different psychological frameworks. 
Lastly, we also experimented with combinations of scaffolds, including synthesizing summary rationales from all scaffolds, but these were less performant (\Cref{sec:apx:combining_scaffolds}).

\section{Using \method with human-written rationales}
\label{sec:human_pilot}
\Cref{table:main_results} demonstrates that plausible yet synthesized LM rationales effectively improve user personas.
However, we hypothesize that the observed improvements stem not from the LM-generated rationales themselves, but from the additional context they provide, enabling more accurate generalization for each user.
To test this, we conducted a pilot study using a subset of 30 questions from \opinionqa. 
We recruited 100 users to respond to the subset of questions, collected their demographics and asked them to provide rationales for the first nine responses. 
These rationales were unconstrained, allowing participants to explain their judgments based on personal experiences, beliefs, or personality traits without any imposed structure. 
The remaining 21 questions were used for evaluation.
Details are in \Cref{sec:apx:human_pilot}.

\Cref{table:human_pilot_results} reports results on this subset, with the inclusion of \humanwritten using \gptfour.
\humanwritten rationales outperform all baselines and \method variants, with the exception of \primalbeliefs, where the difference is not statistically significant. 
This highlights that while \method generates synthetic rationales, plausible and carefully selected scaffolds hold similar predictive power to \humanwritten rationales.
\section{Discussion}

\begin{figure}[!t]
    \centering
    \includegraphics[width=\linewidth]{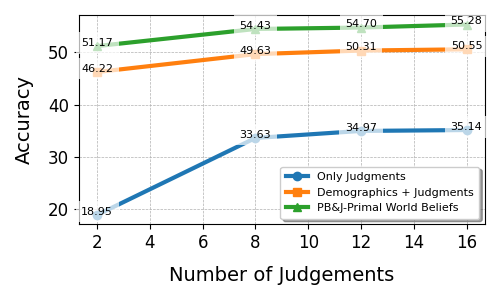}
    \caption{\textbf{Performance as a function of the number of user judgments:} \method outperforms baselines across all settings, providing substantial gains even with minimal judgments. All results use \gptfour.}
    \label{fig:perf_vs_judgements}
\end{figure}

\paragraph{LM-generated rationales help even with a limited budget of judgments.} 
Unlike demographics, user judgments take time to collect, making it crucial to assess personas with limited data (\Cref{sec:experiment}).
As shown in \Cref{fig:perf_vs_judgements}, \method with \primalbeliefs outperforms both baselines (\onlyjudge and \hwang 
) across varying \# of judgments. 
Even with just two judgments, \method surpasses baselines far more, highlighting the value of LM-generated rationales in low-data settings.
As the number of judgments increases, the performance of all methods improves. 
However, the rate of improvement for all diminishes after 8 judgments.
This indicates that while all methods benefit from additional user judgments, \method maximizes its potential gains earlier due to the contextual richness provided by LM-generated rationales.

\paragraph{\method improves performance across demographics.}

\begin{figure}
    \centering
    \includegraphics[width=\linewidth]{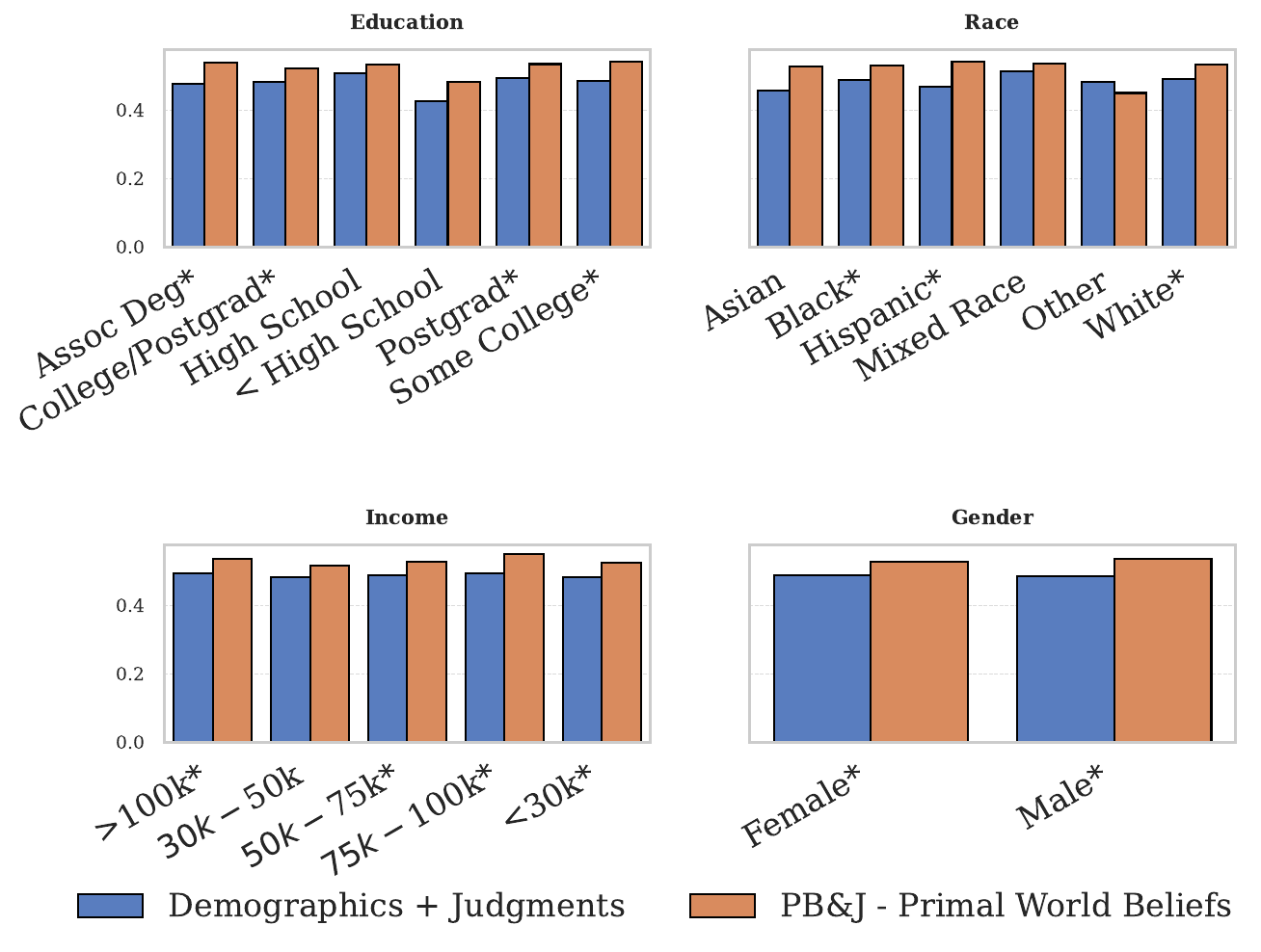}
    \caption{\textbf{\method's improvements over \hwang across education, race, income, and gender:} Subgroups marked with * indicate significant improvements (p < 0.05). All results use \gptfour.}
    \label{fig:subpopulation}
\end{figure}
A robust personalization method should improve performance across diverse user groups rather than relying on gains from specific demographics. 
To assess this, we compare \method (with \primalbeliefs scaffolds) to \hwang across education, race, income, and gender in \Cref{fig:subpopulation} (splits across all demographics in \Cref{fig:subpopulation_all}).
\method consistently outperforms the baseline across all demographics. 
While gains are notable for users with postgraduate education and those identifying as Asian or Mixed Race, improvements extend across all education levels, racial groups, income brackets, and genders. 
By incorporating scaffolded rationales, \method improves with diverse user perspectives, demonstrating broad effectiveness with synthetic rationales.

\paragraph{\method improves performance {\it without} demographics.}
In cases where user demographics are unavailable or where the risks of bias outweigh the utility of demographic information, \method can improve performance of personalization from judgments alone. Rationalizing \onlyjudge with \primalbeliefs results in an 11.64\% absolute improvement on OpinionQA and 1.1\% on MovieLens (GPT-4). While still under-performing demographic-based methods, this result indicates that deeper reasoning about user behavior alone may obviate the need for coarse and potentially biasing demographic information.

\paragraph{Effect of reasoning length on performance.} 
While variants of \method improve over default reasoning offered by $\hwang_{CoT}$, we investigate where this improvement comes from, with reasoning length as our control.
On the pilot subset reported in \Cref{sec:human_pilot}, we observe that \humanwritten rationales are much shorter in length (40.73 $\pm$ 25.66 tokens), as compared to chain-of-thought rationales generated by $\hwang_{CoT}$ (59.70 $\pm$ 16.90 tokens) and $\method_{\primalbeliefs}$ rationales (124.12 $\pm$ 38.16 tokens); however, \humanwritten rationales outperform both these variants. 
Across all \method variants in \Cref{sec:human_pilot}, we observe that the correlation between the accuracy and length of rationales (in terms of tokens) for corresponding users is negligible (Pearson's $r$=0.03).
This suggests that performance gains are not simply a result of longer rationales, but stem from the quality and relevance of the information provided in the rationales.
\section{Related Work}

\paragraph{Personalizing LM.}
Recent works have used LM personas to simulate behavior in psycholinguistic and other social science experiments \cite{aher2023usinglargelanguagemodels, Karra2022EstimatingTP,Filippas2023LargeLM, Argyle_2023}.
Specifically, the use of LMs to simulate user responses to surveys, using existing user information like demographic background has been gaining increasing attention \cite{santurkar2023opinionslanguagemodelsreflect, hwang-etal-2023-aligning, durmus2024measuringrepresentationsubjectiveglobal, chuang2024simulatingopiniondynamicsnetworks, do2024choirecharacterizingpredictinghuman, sun2025personadbefficientlargelanguage, moore-etal-2024-large, Dillion2023CanAL, Tjuatja2023DoLE, balepur2025boatdoesfloatimproving}
Recently, that  attention has shifted more towards synthetically augmenting persona information \cite{moon2024virtualpersonaslanguagemodels, Simmons2022MoralML} or completely synthesizing personas generated from seed human data \cite{park2022socialsimulacracreatingpopulated, park2024generativeagentsimulations1000, Park2023GenerativeAI, ge2024scalingsyntheticdatacreation}.

\paragraph{Psychology and Personas.}
Psychologists have long investigated how different sociological and psychological aspects influence personality \cite{mcadams1993stories, Bruner1991TheNC, Pennebaker1999LinguisticSL}.
Recently, researchers have focused on eliciting psychological markers for evaluating LMs  \cite{Hilliard2024ElicitingPT,Karra2022AIPE,serapiogarcía2023personalitytraitslargelanguage}.
Very few works actually incorporate these principles in an LM persona itself \cite{moon2024virtualpersonaslanguagemodels, park2024generativeagentsimulations1000}.

\paragraph{Reasoning and Rationalization.}

Previous works have focused on generating reasoning chains or rationales by either prompting language models \cite{wei2023chainofthoughtpromptingelicitsreasoning,saha2023languagemodelsteachweaker} or learning to generate rationales by fine-tuning on such data \cite{wiegreffe2022measuringassociationlabelsfreetext, ramnath2024tailoringselfrationalizersmultirewarddistillation}.
Recently, there has been a surge in built-in reasoning capabilities in language models via extensive steps \cite{zelikman2022starbootstrappingreasoningreasoning, deepseekai2025deepseekr1incentivizingreasoningcapability}.
\section{Conclusion}

This work introduces \method, a framework that improves LM personas by incorporating plausible, yet synthetic rationales to explain user judgments.
By leveraging psychological scaffolds, \method improves LM persona accuracy across diverse opinion prediction and preference modeling tasks, while performing at best close to human-written rationales. 
Additionally, \method remains effective even with limited user judgments, highlighting its potential for real-world personalization applications, with limited user history. 
\section*{Limitations}
\label{sec:limitations}
While \method improves personas through plausible yet synthetic LM-generated rationales, it relies solely on zero-shot prompting for both rationale generation and downstream predictions. 
While this allows for flexibility and adaptability across users, it may not fully capture the complexity and depth of individual reasoning.
Fine-tuning LMs on human-written rationales could further improve personalization by enabling models to learn user-specific patterns rather than relying solely on generated rationales.
Additionally, while these rationales improve performance, we cannot validate their fidelity to actual user reasoning, as no ground-truth rationales are available. 
This limitation is inherent to synthetic-data based persona modeling \cite{moon2024virtualpersonaslanguagemodels, park2024generativeagentsimulations1000}, where plausible explanations generated by LMs may align well with observed user behavior but not necessarily reflect the true underlying motivations. 
However, we emphasize and caution that synthesized rationales can be \textit{a plausible} reason for a judgment, but may not be the exact reason used by a user.
We provide further analyses about this in \Cref{sec:apx:expl_analysis}.
\section*{Ethics Statement}

Our study primarily evaluates \method on U.S.-based user populations, as both \opinionqa and our human pilot study consist of participants located in the United States. 
Our study was conducted under the guidance of an ethics review board.
Additionally, the subset of users selected from \movielens also resides in the U.S. While this ensures consistency in evaluation, it limits the generalizability of our findings to more diverse global populations. 
Future work should explore the effectiveness of \method across different cultural and linguistic contexts to ensure broader applicability.
Since our datasets involve personal judgments on opinion-based questions, some generated responses may reflect viewpoints that could be offensive or controversial.
While we do not directly intervene in the LMs’ generation of rationales, it is crucial to recognize that models can inherit biases present in both training data and user-generated inputs.
Finally, as with any system that models human behavior, there are concerns around user privacy. 
While our work does not use real user data beyond voluntary survey responses, deploying such approaches in real-world settings would require careful consideration of data collection practices, consent mechanisms, and safeguards against potential misuse.
\section*{Acknowledgments}
We thank anonymous reviewers and lab members at Apple and USC NLP for their feedback on this work.
This work was also supported in part by the National Science Foundation under grant IIS-2403437, the Simons Foundation, and the Allen Institute for AI.
Any opinions, findings, conclusions or recommendations expressed in this material are those of the author(s) and do not necessarily reflect the views of the National Science Foundation.
This work was partially done while B. Joshi was at Apple. and S. Swayamdipta was a visitor at the Simons Institute for the Theory of Computing.
B. Joshi was also supported by the Apple Scholars in AL/ML PhD Fellowship.

\bibliography{acl_latex}

\appendix

\section{More on Psychological Scaffolds}
\label{sec:apx:psych_scaffolds}

A rich body of research has extended, debated, and validated each of the psychological scaffolds presented earlier.
Among these, \primalbeliefs stands out as the most linguistically motivated framework, which may explain its strong performance in our experiments. 
As described in \cite{clifton2019primalworldbeliefs}, researchers analyzed historical texts and over 80K tweets using topic modeling and concept extraction to identify statements about how people perceive the world. 
These statements were then categorized through expert coding, consultation with social scientists, and discussions with religious focus groups, leading to the identification of 26 fundamental ``primal world beliefs".
These beliefs encapsulate deep-seated assumptions individuals hold about the world, such as whether the world is inherently safe or dangerous, simple or complex, and abundant or limited. 
Expanding on this work, \cite{Vu_Giorgi_Clifton_Balasubramanian_Schwartz_2022} introduced a Latent Beliefs Model, which leverages transformer-based embeddings and a modified GPT-2 model to automatically infer latent dimensions of human beliefs from social media text. 
This data-driven discovery of worldviews underscores the linguistic basis of primal beliefs and their connection to naturally occurring human rationales.

Another well-established framework in social psychology is \shwartz \cite{shwartz1992test,schwartz2012refining}. 
This theory posits that human values, that are fundamental guiding principles, are organized along universal motivational dimensions that drive behavior. 
Schwartz identifies ten broad value categories, such as self-direction (independence of thought and action), benevolence (concern for others' welfare), and power (desire for dominance or control). 
These values are structured in a circular model, where adjacent values are more compatible, and opposing values (e.g., security vs. stimulation) tend to be in tension. 
A key aspect of Schwartz’s values is their cross-cultural validation; extensive empirical studies have shown that these value dimensions hold across diverse populations, making them a robust framework for modeling user judgments in an LM setting. 
Unlike primal beliefs, which describe broad worldviews, Schwartz’s values provide a structured way to infer decision-making tendencies and moral considerations, making them particularly useful for understanding user preferences in ethical or societal questions.

Similarly, the \bigfive (also known as the OCEAN model) \cite{goldberg1993bigfive, costa1999five, pennebaker2001patterns, schwartz2013personality} offer a comprehensive framework for describing individual differences in personality. 
The Big Five dimensions—Openness to Experience, Conscientiousness, Extraversion, Agreeableness, and Neuroticism—have been extensively validated through psychometric studies and natural language analysis. 
These traits predict a wide range of behaviors, from political preferences to purchasing decisions, and have been found to correlate with linguistic patterns in social media and personal narratives. 
For instance, individuals high in Openness to Experience tend to use more abstract and imaginative language, while those high in Neuroticism are more likely to express negative emotions. 
Given these correlations, Big Five traits serve as a useful scaffold for generating rationales that reflect personality-driven reasoning processes, such as why a highly conscientious user might favor structured decision-making or why an extraverted user might prioritize social considerations.

Each of these psychological scaffolds offers a unique perspective on human behavior: Primal World Beliefs focus on fundamental assumptions about the world, Schwartz’s values provide a structured way to model decision-making, and the Big Five capture stable personality traits that influence judgment and preference formation. 

\section{\method prompts}
\label{sec:apx_prompts}

\method is primarily an inference-time prompt-based strategy to improve LM personas, without requiring any fine-tuning.
We pick the basic structure of our prompt from~\citet{hwang-etal-2023-aligning}. For all psychological scaffolds, we try multiple variations of prompts with varying levels of instructions, definitions and examples.
Varying instructions for scaffolds does not lead to significant changes in performance.
For example, adding more detailed instructions for \bigfive leads to a small, yet insignificant increase of 0.6 points for the \opinionqa \humanwritten subset.

In \Cref{tab:prompts_rationalize} and \Cref{tab:prompts_predict}, we provide the final prompts that we use to generate scaffolded rationales and predictions from both \gptfour and \mistral, that are used to generate results displayed in \Cref{table:main_results} and \Cref{table:human_pilot_results}.

\section{Dataset Details}
\label{sec:apx_dataset}

\subsection{\opinionqa}

\opinionqa \cite{santurkar2023opinionslanguagemodelsreflect} contains fifteen topics, with multiple questions in each topic. 
Users answer several questions for a topic.
There is no 1:1 correspondence between users in different topics.
To this end, we select 50 users per topic, resulting in 750 unique users.
For each user, we separate 8 answered questions and use them as prior judgments (this number changes when we run evaluations with varying number of judgments \ref{fig:perf_vs_judgements}).
All remaining questions as used as a `test set', out of which we randomly select 10 questions for evaluation. 
All configurations and demographic setup is used similar to \citet{hwang-etal-2023-aligning}.

\subsection{\movielens}

\movielens \cite{harper2015movielens} dataset contains user ratings (on a scale of 1-5) for movies belonging to different genres.
These ratings are time-stamped.
While there are limited user demographic attributes, the dataset contains information about a user's age, occupation, location (zipcode) and gender.
We convert the zipcode of every participant to a string-based city/state location \cite{geopy}.
For curating judgments, we order movie ratings based on their timestamp, and pick the first 8 ratings as judgments in the persona, and from the remaining ratings, sample 10 to be used for evaluation.
The judgments include the movie name and a user's rating. 
We augment the judgments with the movie's synopsis and key cast and crew using the IMDb API \cite{cinemagoer}.

\subsection{Human Pilot Dataset}
\label{sec:apx:human_pilot}

For the human pilot experiment, we recruit $100$ participants from a third-party user study platform called User Research International. 
We selected a subset of 30 questions from \opinionqa belonging to 3 different topics: food, economics and America in 2050.
Participants who consent to the study are requested to answer all $30$ questions, but can refuse any question they want.
For $9$ questions, we also ask participants to provide a free-text rationale justifying their selection.

\section{More examples of \method-generated rationales}
\label{sec:apx:examples}

We present examples of personas constructed by \method.
For given users with demographics, we display a representative judgment provided by the user, and present rationales generated by \method using different scaffolds for the same judgment. 
We add examples in \Cref{tab:example_1}, \Cref{tab:example_2} and \Cref{tab:example_3}.

\begin{table*}[!t]
    \footnotesize
    \centering
    \resizebox{\linewidth}{!}{
    \begin{tabular}{p{\linewidth}}
        \toprule
        \textbf{User Demographics:}\\
        Age: 18-29, Citizenship: Yes, Region: South, Education: Postgraduate, Income: \$75,000-\$100,000, Marital status: Married, Political ideology: Liberal, Political party: Democrat, Race: White, Religion: Nothing in particular, Frequency of religious attendance: Never, Gender: Female\\
        \\
        \textbf{Selected Judgment:}\\
        I am somewhat concerned that global climate change will harm me personally at some point in my lifetime.\\
        \midrule
        \midrule
        \textbf{\method-generated rationales:}\\
        \\
        \textbf{\noscaffold} \\
        I believe that the effects of global climate change are increasingly evident and impactful. The rise in natural disasters, unpredictable weather, food and water shortages, even though they might seem distanced, have the potential to directly affect my lifestyle, health, and well-being. It's a concern not just for the distant future, but for my lifetime as well, hence my concern.\\
        \midrule

        \textbf{\experiences} \\
        As a postgraduate scholar majoring in environmental sciences, I am constantly exposed to alarming research findings on global climate change. Personally, I love outdoor activities like hiking and camping, and it's heartbreaking seeing the devastation in forests due to wildfires reportedly linked to climate change. This, coupled with evidence of rising sea levels and increased frequency of extreme weather events, propels my concern about how this can directly impact me and our planet.\\
        \midrule

        \textbf{\bigfive} \\
        My concern about global climate change affecting me personally stems from my high level of Conscientiousness. As someone responsible and oriented toward future planning, I understand the significant risks and consequences posed by climate change. Moreover, my Openness to experience allows me to accept and internalize scientific evidence and reports that underline the severity of the issue. My slight Neuroticism also plays a role, making me more prone to worry about potential negative outcomes, hence my concern about personal harm due to global climate change. \\
        \midrule

        \textbf{\shwartz} \\
        This opinion is grounded in a few of Shwartz' basic human values. Primarily, it expresses my value of Universalism, as I am aware of the interconnections and mutual dependencies in our global ecosystem and am concerned for the welfare of nature. It also links to Security, as the threat of climate change disrupts the stability of my personal life and the wider community. Lastly, the opinion slightly touches on Self-direction. My awareness and concern reflect my informed understanding and independent thought on global matters.\\ 
        \midrule

        \textbf{\primalbeliefs} \\
        This viewpoint aligns with my low Safe World Belief, as one can view global climate change as a significant and imminent threat to stability and personal safety. Despite the advancements in technology and modern conveniences, our safety is still compromised by environmental degradation. Furthermore, my high Enticing World Belief encourages me to explore and understand more about this global issue, its impacts, and the ways to mitigate it. However, my low Alive World Belief also reinforces this concern as climate change does not seem to occur with comprehensible intent or purpose, instead contributing to an unpredictable threat to my life.\\
        \bottomrule
    \end{tabular}}
    \caption{Example 1}
    \label{tab:example_1}
\end{table*}

\begin{table*}[!t]
    \footnotesize
    \centering
    \resizebox{\linewidth}{!}{
    \begin{tabular}{p{\linewidth}}
        \toprule
        \textbf{User Demographics:}\\
         Age: 65+,Citizenship: Yes, Region: Northeast, Education: High school graduate, Income: \$100,000 or more, Marital status: Divorced, Political ideology: Very conservative, Political party: Other, Race: White, Religion: Protestant, Frequency of religious attendance: Never, Gender: Female\\
        \\
        \textbf{Selected Judgment:}\\
        I feel no pressure from family members to marry my partner.\\
        \midrule
        \midrule
        \textbf{\method-generated rationales:}\\
        \\
        \textbf{\noscaffold} \\
        I believe in independence and making my own decisions. Also, having been through a divorce, my family understands and respects my perspective about not wanting to remarry.\\
        \midrule

        \textbf{\experiences} \\
        After my divorce, my family has been exceptionally supportive of my decisions regarding my love life. With my grown children doing well and leading their own lives, they are more focused on my happiness than societal expectations. They have always wanted me to be comfortable and happy, rather than rush into any decisions. That's why I haven't felt any pressure from them to marry my current partner.\\
        \midrule

        \textbf{\bigfive} \\
        As a very conservative person who is also divorced, I've learned the importance of self-reliance and making decisions guided by my own perceptions and experiences (relating to the trait of conscientiousness). My family respects my conservative beliefs and my decision not to remarry, thereby causing no pressure for me to marry my partner. Furthermore, my trait of extroversion makes me capable of effectively communicating my decisions and viewpoints to my family, reinforcing their understanding and respect for my personal choices. \\
        \midrule

        \textbf{\shwartz} \\
        This opinion is rooted in the values of Self-direction and Achievement. My high-income bracket has allowed me a certain degree of independence and control over my life choices, which in turn alleviates any external pressure from family to marry my partner. Moreover, being of a mature age, I have lived through the social norms and expectations often imposed on younger generations. Hence, my achievements in life so far have given me the confidence and autonomy to make my own decisions.\\ 
        \midrule

        \textbf{\primalbeliefs} \\
        As someone with a very high Safe World Belief, I live in a cooperative and comforting environment where my decisions are respected. This harmonious setting also allows for open conversation about my personal choices, including my decision to stay unmarried to my partner. Thus, my family members do not pressure me to marry.\\
        \bottomrule
    \end{tabular}}
    \caption{Example 2}
    \label{tab:example_2}
\end{table*}

\begin{table*}[!t]
    \footnotesize
    \centering
    \resizebox{\linewidth}{!}{
    \begin{tabular}{p{\linewidth}}
        \toprule
        \textbf{User Demographics:}\\
        Age: 30-49, Citizenship: No, Region: South, Education: College graduate/some postgrad, Income: \$50,000-\$75,000, Marital status: Married, Political ideology: Conservative, Political party: Independent, Race: Black, Religion: Protestant, Frequency of religious attendance: More than once a week, Gender: Male\\
        \\
        \textbf{Selected Judgment:}\\
        The promotion of racial and ethnic diversity in the workplace is somewhat important for companies and organizations.\\
        \midrule
        \midrule
        \textbf{\method-generated rationales:}\\
        \\
        \textbf{\noscaffold} \\
        While I believe in the importance of individual skill and merit, I also acknowledge the role racial and ethnic diversity plays in fostering a broad range of perspectives in the workplace. This, in turn, can drive innovation, create a more inclusive environment, and better represent the diverse markets that today's companies and organizations serve. Nevertheless, I don't regard it as the most crucial factor for success. Hiring should always focus primarily on capability and qualifications.\\
        \midrule

        \textbf{\experiences} \\
        As someone who has spent a significant portion of my career working in a corporate setting, I have observed firsthand how diversity contributes to a more innovative and productive work environment. Companies with a wide range of perspectives—from people of all races, ethnicities, and backgrounds—tend to solve problems more creatively and be more adaptable in the face of change. While I believe many factors play into a business's success, I have come to appreciate that variety in cultural background and experience certainly enriches the workplace and can provide a competitive edge.\\
        \midrule

        \textbf{\bigfive} \\
        With my level of conscientiousness, I believe that recruiting a workforce that is diverse allows companies and organizations to draw upon a broader range of insights, skills, and experiences, thereby enhancing their productivity and competitiveness. My openness to experience influences my belief that this diversity brings in new and varied perspectives which leads to innovative problem solving. However, it's only 'somewhat' important because I believe merit should be the primary determining factor for hiring and promotions, rather than meeting a quota system.\\
        \midrule

        \textbf{\shwartz} \\
        This opinion aligns with my value of Universalism, as promoting racial and ethnic diversity in the workplace fosters an environment of tolerance and respect for all people, regardless of their background. Additionally, from a Power perspective, a diverse workplace may offer more opportunities for differing perspectives, which could yield more balanced and inclusive decision-making processes, thereby amplifying social status and control over resources. Furthermore, such diversity contributes to Achievement by elevating a organization's competency and credibility in an increasingly globalized world.\\ 
        \midrule

        \textbf{\primalbeliefs} \\
        As someone who has spent a significant portion of my career working in a corporate setting, I have observed firsthand how diversity contributes to a more innovative and productive work environment. Companies with a wide range of perspectives—from people of all races, ethnicities, and backgrounds—tend to solve problems more creatively and be more adaptable in the face of change. While I believe many factors play into a business's success, I have come to appreciate that variety in cultural background and experience certainly enriches the workplace and can provide a competitive edge.\\
        \bottomrule
    \end{tabular}}
    \caption{Example 3}
    \label{tab:example_3}
\end{table*}

\section{Significance Tests}
\label{sec:apx:significance}
\begin{table*}[!t]
\centering
\resizebox{\textwidth}{!}{%
\begin{tabular}{lll|l|l|l}
\toprule
\textbf{Dataset} & \textbf{Model} & \textbf{Best Baseline} & \textbf{Approach} & \textbf{User-wise Significance} & \textbf{Question-wise Significance} \\
\midrule
\multirow{10}{*}{\opinionqa} & \multirow{5}{*}{\gptfour} & \multirow{5}{*}{\hwang} 
& \noscaffold   & statistic=-8.9082, $p$=1.94e-18 & statistic=-9.0335, $p$=1.04e-19 \\
&&& \experiences   & statistic=-9.8689, $p$=5.57e-22 & statistic=-9.8824, $p$=3.42e-23 \\
&&& \bigfive      & statistic=-9.1648, $p$=2.34e-19 & statistic=-9.0564, $p$=8.47e-20 \\
&&& \shwartz      & statistic=-9.0304, $p$=7.14e-19 & statistic=-8.8108, $p$=7.62e-19 \\
&&& \primalbeliefs & statistic=-9.9742, $p$=2.19e-22 & statistic=-9.8010, $p$=7.61e-23 \\
\cmidrule{2-6}
& \multirow{5}{*}{\mistral} & \multirow{5}{*}{\hwang}
& \noscaffold     & statistic=-6.1466, $p$=6.43e-10 & statistic=-6.1429, $p$=4.26e-10 \\
&&& \experiences   & statistic=-6.8153, $p$=9.68e-12 & statistic=-7.2184, $p$=2.89e-13 \\
&&& \bigfive     & statistic=-4.4573, $p$=4.78e-06 & statistic=-4.6148, $p$=2.00e-06 \\
&&& \shwartz      & statistic=-3.5916, $p$=1.75e-04 & statistic=-3.7695, $p$=8.24e-05 \\
&&& \primalbeliefs & statistic=-3.9861, $p$=3.69e-05 & statistic=-4.1223, $p$=1.90e-05 \\
\midrule
\multirow{6}{*}{\movielens} & \multirow{3}{*}{\gptfour} & \multirow{3}{*}{\onlydemo} 
& \bigfive       & statistic=-0.2627, $p$=0.397   & statistic=-0.3260, $p$=0.372 \\
&&& \shwartz     & statistic=-2.6476, $p$=0.00472 & statistic=-3.1586, $p$=0.00082 \\
&&& \primalbeliefs & statistic=-1.7573, $p$=0.04098 & statistic=-1.8790, $p$=0.03027 \\
\cmidrule{2-6}
& \multirow{3}{*}{\mistral} & \multirow{3}{*}{\onlydemo} 
& \bigfive       & statistic=-3.0672, $p$=0.00139 & statistic=-3.5543, $p$=0.00020 \\
&&& \shwartz     & statistic=-1.8034, $p$=0.03718 & statistic=-2.1552, $p$=0.01569 \\
&&& \primalbeliefs & statistic=-3.8985, $p$=8.81e-05 & statistic=-4.1427, $p$=1.86e-05 \\
\midrule
\multirow{6}{*}{Human Pilot} & \multirow{6}{*}{\gptfour} & \multirow{6}{*}{\hwang} 
& \noscaffold     & statistic=-4.9186, $p$=1.74e-06 & statistic=-5.5086, $p$=2.03e-08 \\
&&& \experiences   & statistic=-2.4163, $p$=0.00876 & statistic=-2.9501, $p$=0.00161 \\
&&& \bigfive      & statistic=-3.0124, $p$=0.00165 & statistic=-3.5183, $p$=0.00022 \\
&&& \shwartz      & statistic=-3.4111, $p$=0.00047 & statistic=-4.0207, $p$=3.00e-05 \\
&&& \primalbeliefs & statistic=-7.2335, $p$=5.05e-11 & statistic=-7.8278, $p$=3.90e-15 \\
&&& Human Pilot   & statistic=-8.7457, $p$=2.96e-14 & statistic=-9.2996, $p$=1.71e-20 \\
\bottomrule
\end{tabular}}
\caption{\textbf{Statistical significance of \method's improvements for both \gptfour and \mistral:} Results from one-sided paired t-tests comparing \method variants to the best-performing baselines. We report both user-wise and question-wise significance.
}
\label{tab:apx:significance}
\end{table*}

We conduct statistical tests to assess whether the improvements of \method over the best-performing baseline are statistically significant.
\Cref{tab:apx:significance} presents significance test results, comparing each \method variant with the strongest baseline in its respective setting.
We use a one-tailed independent t-test \cite{student1908probable} to evaluate the null hypothesis that \method does not provide a significant improvement over the baseline. 
To ensure robustness, we compute significance in two ways. 
The first approach, user-wise significance, examines whether \method improves performance on a per-user basis, assessing whether predictions for individual users show meaningful gains. 
The second, question-wise significance, evaluates improvements across all instances of a user, aggregating performance over multiple questions answered by the same user.
For each comparison, we report the test statistic and p-value in \Cref{tab:apx:significance}.

\section{Combining Psychological Scaffolds}
\label{sec:apx:combining_scaffolds}
\begin{table}[!htbp]
\centering
\resizebox{\linewidth}{!}{
\begin{tabular}{lcc}
\toprule
\textbf{Approach}  & \textbf{Accuracy} \\
\midrule
$\method_{\primalbeliefs+\shwartz}$ & 45.19 \small{$\pm$ 10.74} \\
$\method_{\experiences+\bigfive}$ & 42.67 \small{$\pm$ 11.45} \\
$\method_{\textsc{ConcatAll}}$ & 41.40 \small{$\pm$ 11.21} \\
$\method_{\textsc{CombineAllToOne}}$ & 44.28 \small{$\pm$ 11.20} \\
\bottomrule
\end{tabular}}
\caption{\textbf{Combining Scaffolded Rationales:} Given that scaffolds are key to improve LM persona, we ask \textit{to what extent can scaffolded rationales help?} We experiment with four varying settings where scaffolded rationales are combined.}
\label{table:combined_scaffolds}
\end{table}

Given that psychological scaffold-based rationales help improve LM personas, we also investigated settings where we combined rationales from multiple scaffolds for a user.
On the \opinionqa subset containing the \humanwritten rationales, we concatenate the top two (\primalbeliefs and \shwartz), bottom two (\experiences and \bigfive) and all rationales (\textsc{ConcatAll}) as context for the LM persona.
This leads to subpar performance; we posit that adding rationales from all scaffolds is too noisy for the LM to be able to select reasonable justifications to support user judgments.
In order to mitigate this, inspired by Self-Consistency~\cite{wang2023selfconsistencyimproveschainthought}, we consolidate rationales from all scaffolds into a single rationale (\textsc{CombineAllToOne}).
An additional LM is used for this post hoc processing, where the LM is provided the following instructions: ``\textit{For a given judgment, you will be provided multiple rationales for why this person holds this judgment. Your job is to consolidate these rationales into one concise rationale. If the rationales are not consistent with each other or present diverging perspectives, you are allowed to pick a perspective, or also allowed to keep multiple perspectives for that judgment, based on what you think best reflects the person.}''
While this improves over the \textsc{ConcatAll} setting, this is still not enough signal for the LM persona, unlike single scaffolds like \primalbeliefs.

\begin{table*}[!t]
    \footnotesize
    \centering
    \resizebox{\linewidth}{!}{
    \begin{tabular}{p{\linewidth}}
        \toprule
        \textbf{System Message:}\\
        A person can be described as follows: \\
        \\
        <demographic information>\\
        \\
        The person has the following judgements:
        
        1. <judgement\_1>\\
        2. <judgement\_2>\\
        3. <judgement\_3>\\
        \midrule
        \midrule
        \textbf{User Message:}\\
        \\
        \textbf{\noscaffold} \\
        For a given judgement, what would be a reasonable explanation that the person would provide for holding that judgement?\\
        \\
        Judgement: <judgement\_i>\\
        \midrule

        \textbf{\experiences} \\
        For a given judgement, what would be a reasonable explanation that the person would provide for holding that judgement? The explanation should contain a specific experience or personality trait (for example, fill in details of the university, or food, or any other detail, that the person would use to better explain their judgment).\\
        \\        
        Judgement: <judgement\_i> \\ 
        \midrule

        \textbf{\bigfive} \\
        For a given judgement, what would be a brief, reasonable explanation that the person would provide for holding that judgement? The explanation should be grounded in the big five personality traits listed below:\\
        \\
        1. Openness to experience (includes aspects such as intellectual curiosity and creative imagination)\\
        2. Conscientiousness (organization, productiveness, responsibility)\\
        3. Extroversion (sociability, assertiveness; its opposite is Introversion)\\
        4. Agreeableness (compassion, respectfulness, trust in others)\\
        5. Neuroticism (tendencies toward anxiety and depression)\\
        \\
        Judgement: <judgement\_i> \\
        \midrule

        \textbf{\shwartz} \\
        For a given judgement, what would be a brief, reasonable explanation that the person would provide for holding that judgement? The explanation should be grounded in the Shwartz Theory of Basic Human Values listed below: \\
        \\
        1. Power: Refers to the pursuit of social status, dominance, and control over people and resources.\\
        2. Achievement: Personal pursuit of success, demonstrating competence according to social standards.\\
        3. Hedonism: Pursuit of pleasure, enjoyment, and sensory and emotional gratification.\\
        4. Stimulation: Seeks novelty and challenge in life, valuing excitement, variety, and adventure.\\
        5. Self-direction: Independent thought and action — choosing, creating, and exploring.\\
        6. Universalism: Understanding, appreciation, tolerance, and protection for the welfare of all people and nature.\\
        7. Benevolence: Preserving and enhancing the welfare of those with whom one is in frequent personal contact (the ‘in-group’).\\
        8. Tradition: Respect, commitment, and acceptance of the customs and ideas that traditional culture or religion provide the self.\\
        9. Conformity: Restraint of actions, inclinations, and impulses likely to upset or harm others and violate social expectations or norms.\\
        10. Security: Safety, harmony, and stability of society, relationships, and the self. \\
        \\
        Judgement: <judgement\_i> \\ 
        \midrule

        \textbf{\primalbeliefs} \\
        For a given judgement, what would be a brief, reasonable explanation that the person would provide for holding that judgement? The explanation should be grounded in the three primal world beliefs listed below:\\
        \\  
        1. Safe World Belief: Those low on Safe see a Hobbesian world defined by misery, decay, scarcity, brutality, and dangers of all sorts. Those high on Safe see a world of cooperation, comfort, stability, and few threats.\\
        2. Enticing World Belief: Those low on Enticing inhabit dull and ugly worlds where exploration offers low return on investment. Those high on Enticing inhabit an irresistibly fascinating reality.\\
        3. Alive World Belief: Those low on Alive inhabit inanimate, mechanical worlds without awareness or intent. Those high on Alive sense that everything happens for a purpose and are thus sensitive to those purposes.\\
        \\
        Judgement: <judgement\_i>\\
        \bottomrule
    \end{tabular}}
    \caption{\textbf{Prompts used to generate rationales for judgments:} We use a common system message that includes a user's demographic information and all prior judgments held by the user. The user message then includes scaffold specific instructions $\psi$ to generate rationales for a specific judgment.}
    \label{tab:prompts_rationalize}
\end{table*}

\begin{table*}[!t]
    \footnotesize
    \centering
    \resizebox{\linewidth}{!}{
    \begin{tabular}{p{\linewidth}}
        \toprule
        \textbf{System Message:}\\
        You are the following person: \\
        \\
        <demographic information>\\
        \\
        You have the following opinions:

        1. <judgement\_1>+<rationale\_1>\\
        2. <judgement\_2>+<rationale\_2>\\
        3. <judgement\_3>+<rationale\_3>\\
        \midrule
        \midrule
        \textbf{User Message:}\\
        \\
        Based on your demographic and opinion information above, which answer would you select for the question shown below?\\
        \\
        Question: <question>\\
        Answer choices: <choice>\\    
        \midrule
        \textbf{User Message with Chain of Thought:}\\
        \\
        Based on the above list of opinions and the demographic information, what would you choose for the question shown below?
        Provide your answer in the following format - "Reason: <reason>, Answer: <answer>". Only answer amongst the provided options, nothing else. Do not abstain from answering. \\
        \\
        Question: <question>\\
        Answer choices: <choice>\\    
        
        \bottomrule
    \end{tabular}}
    \caption{\textbf{Prompts used to predict answers, given a persona:} We use a common system message that includes a user's demographic information and all prior judgments, along with generated rationales. The user message then includes the exact question and answer choices, with or without rationales.}
    \label{tab:prompts_predict}
\end{table*}

\section{Model Details}
\begin{table*}[!t]
    \centering
    \resizebox{0.8\linewidth}{!}{
    \begin{tabular}{cll}
        \toprule
        \textbf{Config} & \textbf{\gptfour} & \textbf{\mistral}\\
        \midrule
        \multirow{2}{*}{model} 
        & \textbf{\gptfour-0613} & \textbf{\mistral 0.2 Instruct} \\
        & Number of parameters: Unknown
        & Number of parameters: 7 billion \\
        \midrule
        \midrule
        \textbf{Rationale Generation} & & \\
        new\_tokens & 256 & 256\\
        temperature & 1 & 1\\
        seed & 6 & 6\\
        GPU & N/A, openai api call & 3 A100\\
        Inferring time & 2 hours & 1 hour\\
        \midrule
        \textbf{Answer Generation} & & \\
        new\_tokens & 10 & 275\\
        temperature & 0 & 0\\
        seed & 6 & 6\\
        GPU & N/A, openai api call & 3 A100\\
        Inferring time & 2 hours & 1 hour\\
        \bottomrule
        \bottomrule
    \end{tabular}
    }
    \caption{Model Configurations for Rationale and Answer Generation}
    \label{tab:rationale_generation_config}
\end{table*}

We provide model configurations for both \gptfour and \mistral for rationale generation and answer generation steps in \Cref{tab:rationale_generation_config}.\
\gptfour is under proprietary license, and \mistral is subject to the Apache 2.0 license.

\section{Additional Discussion}

\subsection{Does Point-of-View (PoV) matter while building LM personas?}
\begin{table}[!h]
\centering
\resizebox{\columnwidth}{!}{
\begin{tabular}{lccc}
\toprule
\textbf{Approach} & \textbf{Rationale PoV} & \textbf{Answer PoV} & \textbf{Accuracy} \\ 
\midrule
\hwang & -  & first & 32.05\\ 
\hwang  & -  & third & \underline{49.63}\\ 
$\method_{\primalbeliefs}$ & first  & first & \textbf{54.43}\\ 
$\method_{\primalbeliefs}$ & first  & third & 50.42\\ 
$\method_{\primalbeliefs}$ & third  & first & 53.08\\ 
$\method_{\primalbeliefs}$ & third  & third & 51.12\\ 
\bottomrule
\end{tabular}
}
\caption{\textbf{Ablations with different Point-of-Views (PoVs) in \method:} We experiment with different PoVs to prompt answers and rationales, for the best forming baseline, and variant of \method. All experiments on \gptfour.}
\label{table:pov_experiments}
\end{table}

Our analysis examines how the point of view (PoV) in rationale and answer generation affects persona alignment in \method. 
The baseline model benefits from using a \textit{third}-person PoV for answers, suggesting that distancing the model from a subjective stance improves alignment. 
In contrast, \method performs best when both the rationale and answer are generated in \textit{first}-person, reinforcing the effectiveness of maintaining a consistent, personalized perspective. 
Performance declines when either the rationale or answer shifts to \textit{third}-person, indicating that while baselines may benefit from objective framing, \textit{first}-person perspectives enhance persona consistency when paired with rationale-augmented personas.
Therefore, all baselines depicted in \Cref{table:main_results} are prompted in \textit{third}-person and all \method variants are prompted in \textit{first}-person, for both the rationale and answer prompts.



\section{Analysis of LM generated rationales}
\label{sec:apx:expl_analysis}
\begin{table*}[!t]
\centering
\resizebox{\linewidth}{!}{
\begin{tabular}{lcccc}
\toprule
\multicolumn{1}{c}{\textbf{Approach}} & \multicolumn{1}{c}{\textbf{\begin{tabular}[c]{@{}c@{}}\% of rationales \\ that have \\ atleast one \\ demographic\\  mentioned ($\uparrow$)\end{tabular}}} & \multicolumn{1}{c}{\textbf{\begin{tabular}[c]{@{}c@{}}Demographic \\ Keyword \\ Density \\ (\# of demographic \\ keywords / \\ rationale) ($\uparrow$)\end{tabular}}} & \multicolumn{1}{c}{\textbf{\begin{tabular}[c]{@{}c@{}}Avg. diversity \\ of 2-grams\\  in rationales \\ per user ($\downarrow$)\end{tabular}}} & \multicolumn{1}{c}{\textbf{\begin{tabular}[c]{@{}c@{}}Avg. diversity \\ of 2-grams \\ in rationales per \\ judgment ($\downarrow$)\end{tabular}}} \\
\midrule
\midrule
$\method_{\noscaffold}$& 94.66 & 1.26 \small{$\pm$0.50} & 4.15 \small{$\pm$0.12} & 4.18 \small{$\pm$0.34}  \\
$\method_{\experiences}$& 99.25 & 1.57 \small{$\pm$0.47} & 4.66 \small{$\pm$0.11} & 4.64 \small{$\pm$0.29}  \\
$\method_{\bigfive}$ & 94.96 & 1.41 \small{$\pm$0.57} & 4.60 \small{$\pm$0.12} & 4.54 \small{$\pm$0.33}  \\
$\method_{\shwartz}$& 89.03 & 1.14 \small{$\pm$0.54} & 4.61 \small{$\pm$0.11} & 4.51 \small{$\pm$0.36}  \\
$\method_{\primalbeliefs}$& 89.33 & 1.22 \small{$\pm$0.61}  & 4.83 \small{$\pm$0.13} & 4.75 \small{$\pm$0.34}  \\   
\bottomrule
\end{tabular}
}
\caption{\textbf{Analysis of LM-generated explanations across psychological scaffolds, measuring demographic mentions, keyword density, and 2-gram diversity:} Results suggest that strong persona alignment does not necessarily require high demographic reliance or linguistic variation. All analysis on rationales generated by \gptfour.}
\label{table:explanation_analysis}
\end{table*}

\paragraph{Context analysis of LM-generated rationales}
To better understand how different scaffolds influence the content and structure of LM-generated rationales, we analyze the rationales along three key dimensions: demographic reliance, lexical diversity, and structural repetition (\Cref{table:explanation_analysis}). 
The percentage of rationales mentioning demographics and the density of demographic keywords capture how often rationales explicitly reference user identity. 
The average 2-gram diversity per user and per judgment reflects the linguistic variability of rationales, with lower values indicating greater diversity (i.e., less repetition).
This score is also known as self-repetition \cite{salkar-etal-2022-self, shaib2024standardizingmeasurementtextdiversity}.

Our findings reveal distinct patterns across scaffolds. 
Experience-based rationales ($\method_{\experiences}$) exhibit the strongest reliance on demographic cues, with the highest demographic mention rate (99.25\%) and density (1.57), suggesting that user experiences are often framed explicitly through identity attributes. 
Trait- and value-based scaffolds ($\method_{\bigfive}$, $\method_{\shwartz}$) show less reliance on demographics, producing rationales that generalize beyond identity markers. 
Belief-driven rationales ($\method_{\primalbeliefs}$) are highly structured but more repetitive, showing the lowest lexical diversity across users and judgments. 
Notably, unstructured rationales ($\method_{\noscaffold}$) generate the most varied rationales, highlighting that scaffolding provides consistency at the cost of lexical diversity.
These findings suggest that structured reasoning, rather than demographic emphasis or linguistic diversity, is key to improving persona alignment. 
Scaffolds like \primalbeliefs highlight how value-based rationales can enhance LM personalization without overfitting to surface-level demographic traits.

\paragraph{Comparing of LM-generated rationales with human-written.}
While we emphasize that \method generates plausible yet synthetic rationales that \textit{may} not be the actual reasoning user by the user, we acknowledge in \Cref{sec:limitations} that this is indeed a problem with any synthetic data used in an LM pipeline. 
However, we compare and contrast rationales generated by \method with \humanwritten ones on the \opinionqa \humanwritten subset.
\Cref{tab:pb_vs_human_rationales} presents examples that fall into two broad categories: cases where human and LM rationales reflect similar reasoning, and cases where they diverge. 
In the former, LM rationales often elaborate on the human rationale, mapping it onto structured beliefs such as the Safe or Enticing World. 
In the latter, human rationales sometimes cite specific personal anecdotes or cultural references that are absent in LM generations, which tend to rely on general psychological framing.

Despite this variability, \method still benefits from the structure provided by scaffolds like \primalbeliefs, often yielding rationales that are plausible and useful for prediction, even when the reasoning path differs from that of the original user. 
This highlights both the promise and limitations of structured, synthetic rationalization for modeling human behavior.

\begin{table*}[!t]
    \footnotesize
    \centering
    \resizebox{0.9\linewidth}{!}{
    \begin{tabular}{p{\linewidth}}
        \toprule
        \textbf{Example Type:} Similar Reasoning\\
        \textbf{Selected Judgment:}\\
        The growing number of legal immigrants working in the U.S. does not contribute too much to economic inequality in the country.\\\\
        \textbf{\primalbeliefs rationale:}\\
        \\
        I firmly stand on the opinion that the growing number of legal immigrants in the U.S. does not contribute significantly to economic inequality. Grounded in the Safe World Belief, I perceive the world as one of cooperation where diverse people can coexist harmoniously. Immigrants often fill roles in our workforce that are frequently overlooked, thus contributing to the economy and overall stability of our country.

        From the Enticing World Belief perspective, our world is fascinating and enriched by the presence of individuals of different cultures and skillsets. Legal immigrants bring unique perspectives and cultural norms that can actually enhance our socioeconomic environment.
        
        Finally, looking through the lens of the Alive World Belief lens, I understand that people immigrate to the U.S. for a better life, not to contribute to economic disparity. They serve a purpose in the economy, thereby contributing to the betterment of society as a whole. \textit{They are not the ones who create or increase economic inequality. Instead, it's systemic policies and structures that need addressing.}\\\\
        \textbf{\humanwritten rationale:}\\
        \\
        With or without them, capitalism is designed to make economic inequality.
        \\
        \midrule
        \textbf{Example Type:} Similar Reasoning\\
        \textbf{Selected Judgment:}\\
        All in all, I favor the use of animals in scientific research.\\\\
        \textbf{\primalbeliefs rationale:}\\
        \\
        My belief is rooted in the Enticing World belief. I see the world as a place of progress and innovation. The use of animals in scientific research enables us to make strides in understanding diseases and developing treatments. While I acknowledge it's a controversial topic due to ethical considerations, I believe the benefits outweigh the negatives. This kind of research promises great return, potentially leading to medical breakthroughs that can enhance human health and longevity.\\\\
        \textbf{\humanwritten rationale:}\\
        \\
        Although I do not endorse animal cruelty or abuse, I think some form of scientific research using animals is necessary for the greater human good. Even though these animals might suffer or die, they are helping to advance human society.\\
        \midrule
        \textbf{Example Type:} Not Similar Reasoning\\
        \textbf{Selected Judgment:}\\
        Medical treatments today are considered worth the costs because they enable individuals to live longer and enjoy better quality lives.\\\\
        \textbf{\primalbeliefs rationale:}\\
        \\
        I hold this opinion mainly because of my high score on the safe world belief. I believe in a world of cooperation and comfort, where people come together to ensure the wellbeing of everyone. Today's medical treatments, despite their costs, allow individuals to live longer, healthier lives, which contributes to a safer, more comfortable world. I also have a high enticing world belief because medical advancements hold the possibility of new discoveries, further enhancing our quality of life and making the world more enticing. Lastly, my moderate alive world belief stresses the significance of purpose in every happening, and I see the purpose of these treatments in promoting longer and healthier lives.\\\\
        \textbf{\humanwritten rationale:}\\
        \\
        I loved listening to A Prairie Home Companion on NPR. A while back, the host, Garrison Keillor, had a heart valve replacement. Many of his ancestors had just died at his age, but he got to entertain us for another 20 and is still touring at 82.\\
        \midrule
        \textbf{Example Type:} Not Similar Reasoning\\
        \textbf{Selected Judgment:}\\
        Medical treatments today often create as many problems as they solve.\\\\
        \textbf{\primalbeliefs rationale:}\\
        \\
        I believe that we are living in a Safe World, where medical technologies are advancing rapidly. However, sometimes, this progression can be seen as a double-edged sword. While it offers the possibility to treat illnesses, it can also lead to unforeseen side effects and complications, potentially causing more harm than good. This is a result of the complexity of the human body and the inherent risks associated with medical treatments. We must balance the desire for healing with an understanding of the potential risks involved.\\\\
        \textbf{\humanwritten rationale:}\\
        \\
        Over priced medical care\\
        
        \bottomrule
    \end{tabular}}
    \caption{We provide examples of \primalbeliefs and \humanwritten rationales for the same judgment and user. We observe that sometimes, LM-generated rationales reflect the same reasoning conveyed by the human-written ones, however, in cases where the human-written rationales are very niche, it is hard for the LM-generated rationales to mimic the same reasoning.}
    \label{tab:pb_vs_human_rationales}
\end{table*}

\begin{figure*}[!t]
    \centering
    \includegraphics[width=0.8\linewidth]{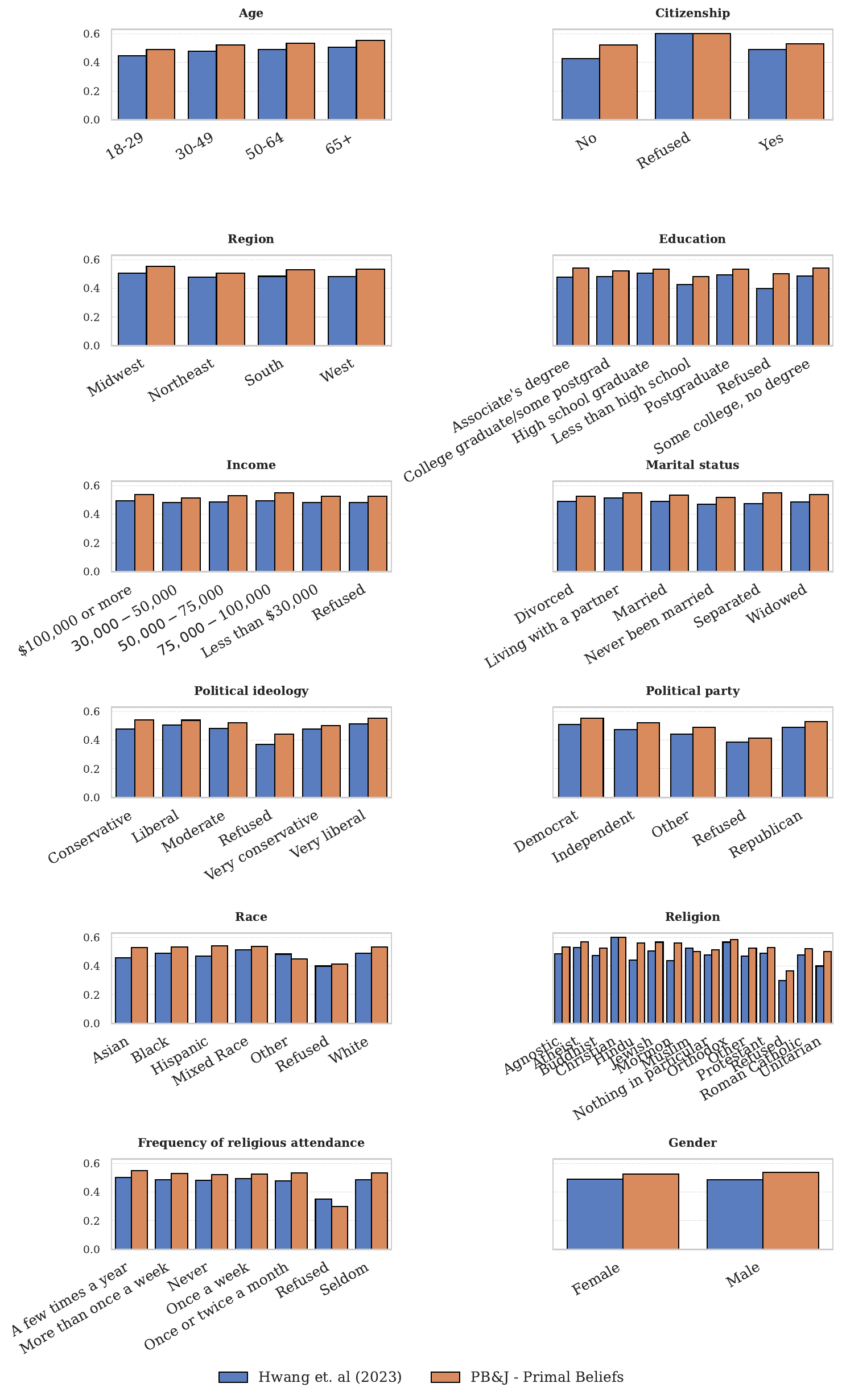}
    \caption{\textbf{\method's improvements over \hwang across all demographics:} Subgroups marked with * indicate significant improvements (p < 0.05). All results use \gptfour.}
    \label{fig:subpopulation_all}
\end{figure*}

\section{Wasserstein Distance Metrics}
\label{sec:apx:wsd}
\begin{table}[!h]
\centering
\resizebox{\linewidth}{!}{
\begin{tabular}{lccc}
\toprule
\textbf{Approach}  & \textbf{\gptfour} & \textbf{\mistral} \\ 
\midrule
\nopersona & 0.95 \small{$\pm$ 0.50} & 2.66 \small{$\pm$ 0.66} \\ 
\onlydemo  & 0.75 \small{$\pm$ 0.42} & 0.97 \small{$\pm$ 0.53} \\
\onlyjudge & 0.83 \small{$\pm$ 0.44} & 1.31 \small{$\pm$ 0.62}\\
\hwang  & 0.81 \small{$\pm$0.45} & 1.03 \small{$\pm$ 0.56}\\ 
$\hwang_{CoT}$  & 0.88 \small{$\pm$0.39} & 0.97 \small{$\pm$ 0.27}  \\ 
\midrule
$\method_{\noscaffold}$  & 1.33 \small{$\pm$ 0.69}  & 1.41 \small{$\pm$ 0.69}\\ 
$\method_{\experiences}$  & 0.99 \small{$\pm$ 0.61} & 1.45 \small{$\pm$ 0.72} \\
$\method_{\bigfive}$ & 0.84 \small{$\pm$ 0.63} & 0.64 \small{$\pm$ 0.31}\\
$\method_{\shwartz}$  & 0.61 \small{$\pm$ 0.33} & 1.14 \small{$\pm$ 0.67}\\
$\method_{\primalbeliefs}$ & 0.64 \small{$\pm$ 0.34} & 0.61 \small{$\pm$ 0.32}\\ 
\bottomrule
\end{tabular}}
\caption{\textbf{Wasserstein Distance Metrics for \movielens}}
\label{table:wsd_movielens}
\end{table}

In addition to accuracy, we also present the Wasserstein Distance (WSD) between predicted ratings and user-provided ratings for the \movielens dataset. 
We present these results in \Cref{table:wsd_movielens}. 
Note that lower values are better for Wasserstein Distance.

\end{document}